\documentclass[letterpaper]{article}

\usepackage{natbib,alifeconf}  

\usepackage{booktabs} 
\usepackage[ruled,vlined]{algorithm2e}
\usepackage{makecell}
\usepackage{amsfonts}  
\usepackage{subfigure}
\usepackage{hyperref}

\title{Quality Evolvability ES: Evolving Individuals With a Distribution of Well Performing and Diverse Offspring}
\author{Adam Katona$^{1}$, Daniel W. Franks$^{1}$ \and James Alfred Walker$^{1}$ \\
\mbox{}\\
$^1$Department of Computer Science, University of York, UK \\
ak1774@york.ac.uk} 

\begin{document}
\maketitle

\begin{abstract}
One of the most important lessons from the success of deep learning is that learned representations tend to perform much better at any task compared to representations we design by hand. Yet evolution of evolvability algorithms, which aim to automatically learn good genetic representations, have received relatively little attention, perhaps because of the large amount of computational power they require. 
The recent method Evolvability ES allows direct selection for evolvability with little computation. However, it can only be used to solve problems where evolvability and task performance are aligned. We propose Quality Evolvability ES, a method that simultaneously optimizes for task performance and evolvability and without this restriction. This is achieved by using nondominated sorting inside the ES update.
Our proposed approach Quality Evolvability has similar motivation to Quality Diversity algorithms, but with some important differences. While Quality Diversity aims to find an archive of diverse and well-performing, but potentially genetically distant individuals, Quality Evolvability aims to find a single individual with a diverse and well-performing distribution of offspring. By doing so Quality Evolvability is forced to discover more evolvable representations. We demonstrate on robotic locomotion control tasks that Quality Evolvability ES, similarly to Quality Diversity methods, can learn faster than objective-based methods and can handle deceptive problems. 


\end{abstract}

\section{Introduction}

Evolution of evolvability is an unintuitive concept. Even the question of whether evolvability evolves at all \citep{pigliucci2008evolvability} is unclear, and there is still debate as to whether evolution of evolvability is caused by natural selection, or is a by-product of other evolutionary mechanisms \citep{pigliucci2008evolvability,payne2019causes}. While this subject remains highly debated among biologists, researchers within the field of evolutionary computation are also increasingly captivated. Evolution of evolvability is desirable because it allows the possibility of accumulating the ability to evolve for a long time, potentially enabling the speed up of future evolution by many orders of magnitude, allowing us to utilize evolution in areas that were not practical before.  

Recent work has enhanced our understanding by examining the evolution of evolvability from a learning theory perspective \citep{watson2016can}, leading to insights that are critical to consider when designing an evolution of evolvability algorithm. The reason it is possible for evolution to increase its ability to evolve in the future is similar to how it is possible for learning to generalize to unseen data \citep{watson2016can}. It requires finding general patterns in past experiences. If some aspect of a genome was useful for generating adaptations in many different past environments, it might be useful in new, unseen environments as well. The job of an algorithm that aims to increase evolvability is to find these general patterns. 

Looking at evolvability from a learning theory perspective, we identify three questions that need to be investigated before we are able to realize the vision of evolution of evolvability. These cover the three basic blocks of any learning process, data containing a general pattern, a model with capacity to learn, and an algorithm to drive learning.

\subsection{Evolvability Data}

The first question is how to provide evolution with sufficiently diverse data that makes it possible for evolvability to evolve. This requires evaluating the evolvability of genes in multiple settings, otherwise, generalization is not possible. This data can come from a single run, but at different stages of evolution (what worked in the past), like in nature. However, this makes evolvability prone to forgetting. There is no way to go back and check if a gene still has the ability to be evolvable in past environments. A more powerful way to provide diverse data is to simultaneously evaluate genomes in a distribution of environments, similarly to the meta learning problem formulation \citep{vilalta2002perspective}. Another key piece of the puzzle might be algorithms that co-evolve environments with agents in order to maximize learning, like the POET algorithm \citep{wang2019paired}.

\subsection{Evolvability Model}
The second question is how to provide evolution with a model which has sufficient capacity to learn evolvability. A powerful source of such capacity is the developmental program, which provides the instructions to translate a genotype into a phenotype. Different generative encodings can result in varying amount of evolvability\cite{tarapore2015evolvability}. With a well-chosen developmental program, evolution is able to turn random genotypic variation into an advantageous distribution of phenotypic variation \citep{watson2016can}. A simple example to demonstrate how such developmental decisions are able to affect evolvability is to imagine how the leg length of an animal is encoded \citep{huizinga2018emergence}. By utilizing symmetry early in development, evolution can become more likely to explore the configurations where both legs are the same length, and avoid exploring the probably poorly performing phenotypes with different left and right leg lengths. 

Additionally, there is another source of learning capacity, which is simply selecting points in the genotype-space with good quality neighborhoods. In the case of direct encoding, this is the only source, since there is no developmental program. The meta learning algorithm MAML \citep{finn2017model} demonstrated the existence of such good quality neighborhoods in the search space of large neural networks. MAML is able to find points in the search space which are a few steps away from good solutions to many previously unseen problems.


\subsection{Evolvability Algorithm}
The third question is how to efficiently select for evolvability. It is not clear under which conditions evolvability can emerge in nature \citep{pigliucci2008evolvability}, whether it requires selection, or if it is a result of unsupervised learning \citep{watson2016can}. In the case of evolutionary computation, we have the ability to directly select for evolvability. There are several existing algorithms that aim to indirectly select for evolvability, and a few that directly select for it. We give an overview of these methods in the Background Section.

\subsection{Contributions}
The aim of this paper is to contribute to the third question by finding algorithms that can directly and effectively select for evolvability. This paper makes the following contributions:
\begin{itemize}
\item We propose Quality Evolvability, an approach that directly selects for evolvability by finding individuals with a diverse and well-performing distribution of offspring, and we present Quality Evolvability ES a method based on Evolvability ES, which is able to select for Quality Evolvability with a low computational cost. 
\item We demonstrate on robotic locomotion tasks, that Quality Evolvability ES is able to outperform the objective based variant of the algorithm, can be applied to problems where evolvability and performance are not aligned, and is able to handle deceptive problems.
\end{itemize}

The main difference between our work, and previous work \citep{mouret2011novelty,lehman2011evolving} which uses nondominated sorting to simultaneously select for diversity and fitness is that our work selects for the expected diversity of the offspring, compared to the diversity of the population. This is possible to do efficiently because of the special properties of ES to estimate gradients of expectations. Our source code is available at: \href{https://github.com/adam-katona/QualityEvolvabilityES}{\textit{https://github.com/adam-katona/QualityEvolvabilityES}}.

\section{Background}

In this paper, we define evolvability as the ability to generate phenotypic variation, a definition used by previous works \citep{mengistu2016evolvability,gajewski2019evolvability}. This variation is measured in certain dimensions of interest defined by the behavioral characterization (BC) function. In this section, we discuss why we chose this evolvability definition and explore other definitions and their relevance to evolutionary computation. We also review existing algorithms that either indirectly or directly select for evolvability.

\subsection{Evolvability Concepts}

There is an evolvability concept concerned with the ability of a population to respond to selection \citep{flatt2005evolutionary}, which mainly depends on the amount of standing genetic variation. This is the evolvability concept selective breeders care about. The term "introducing new blood" refers to introducing alleles to the gene pool which were lost during the domestication process \citep{mccouch2004diversifying}. The population becomes more responsive to selection as we increase the standing genetic variation. We argue that this is not as interesting for evolutionary computation. Standing variation can only explain short term evolvability, for evolvability to be sustained for a long time, a steady supply of variation needs to be generated. Differences in long-term evolvability depend on the ability to generate variation rather than the currently available variation in the population.

The evolvability concepts that are more interesting for evolutionary computation are concerned with the ability to generate phenotypic variation. This kind of evolvability is about the potential to generate variation, whether it is realized or not. There are several different ways to quantify phenotypic variation, we will discuss three of them in the next section, and why they might be interesting for evolutionary computation.


\subsection{Quantifying Evolvability}

The decision on how to quantify evolvability should be based on which kind of evolvability is best suited for generalizing for future environments. We describe three different ways evolvability can be quantified, all of them capturing a different aspect of evolvability.

\subsubsection{Adaptiveness}
The first approach to quantify evolvability is to measure adaptiveness: which is often a measure of both how often offspring have higher fitness than their parents, and the magnitude of the difference in their fitnesses. \citep{wagner1996perspective,altenberg1994evolution} define evolvability as ``the ability of the genetic operator/representation scheme to produce offspring that are fitter than their parents''.

One algorithm which directly selects for adaptiveness is ES-MAML \citep{song2019maml}, an evolutionary version of the meta learning algorithm MAML \citep{finn2017model}. To evolve evolvability that can generalize to new environments, ES-MAML is trained in a set of environments with the expectation that there is a common pattern between the training environments that allows faster adaptation in the unseen test environments.

\subsubsection{Behavioural Diversity}
The second approach to quantify evolvability is to measure the behavioural diversity of the offspring. Diversity can be measured by the variance or entropy of the distribution of behaviours. This approach requires a user defined behaviour characterization (BC) function, which turns some aspect of the agents behaviour into real valued numbers. Examples of what kind of BCs were used in the past include:  final position of the robot \citep{gajewski2019evolvability}, ratio of time each leg of a robot is in contact with the ground \citep{cully2015robots} and the concatenated RAM state for each step for Atari games \citep{conti2017improving}.
Care must be taken to define a BC where achieving diversity requires individuals to develop skills, so that evolution cannot simply exploit the BC function in a similar way to how it can exploit loopholes in not carefully designed fitness functions \citep{lehman2020surprising}. For example, in a closed maze, the only way to reach new places is to learn to navigate and not run into walls. However, if the maze is open on one side, evolution can create individuals that wander outside of the maze, achieving high diversity of final positions without ever learning navigation skills \citep{lehman2011abandoning}. 
Existing algorithms using this definition of evolvability are Evolvability Search \citep{mengistu2016evolvability} and Evolvability ES \citep{gajewski2019evolvability}.

\subsubsection{Innovation}
The third approach is to define evolvability as the ability to generate innovation. \cite{brookfield2001evolution} defines evolvability as ``the proportion of radically different designs created by mutation that are viable and fertile''. However, innovation or ``radically different'' are subjective terms. Furthermore, something that was an innovation in the past, may no longer be considered innovative now. One way to measure innovation is to calculate novelty compared to an archive of previously observed behaviours. This is a similar concept to Novelty Search \citep{lehman2011abandoning}. It is important to note that while novelty can be used to quantify evolvability, Novelty Search does not directly select for evolvability. Evolvability search is interested in discovering individuals which are good at generating novelty, whereas Novelty Search aims to find novel individuals. The hope with this kind of evolvability definition is that if we can find the patterns which made an individual good at generating innovation in the past, maybe these patterns will be useful for generating innovation in the future.  An algorithm that directly selects for this kind of evolvability is Novelty Search ES (NS-ES) \citep{conti2017improving}.

\subsection{Evolvability Algorithms}

There are various existing algorithms which select for evolvability either directly or indirectly.

\subsubsection{Indirect Selection}
There are various mechanisms that produce indirect selection for evolvability \citep{mengistu2016evolvability}. One mechanism is to introduce regular mass extinction events \citep{lehman2015enhancing}, freeing up many niches. Evolvability is indirectly rewarded because lineages that have the ability to radiate to the empty niches faster have a higher chance of surviving the next extinction event. A similar situation can be achieved with constant goal switching \citep{nguyen2015innovation}, which indirectly rewards individuals that developed the ability to adapt between the goals faster. Novelty search is another method that rewards radiating into new niches, and therefore indirectly rewards evolvability. Another mechanism is innovation protection, which protects new genes for a few generations. This protection allows genes that increase evolvability to stay alive long enough to realize the benefits of increased evolvability \citep{stanley2002evolving,risi2019improving}. 

These are all indirect methods, because they do not directly reward evolvability, but create a situation where lineages that are evolvable have some benefit. For this reason, these methods are vulnerable to being exploited by individuals which happen to be novel without being evolvable.

\subsubsection{Direct Selection}
Even though  techniques to increase evolvability have been discussed for several decades, the first technique to directly select for evolvability, Evolvability Search \citep{mengistu2016evolvability}, was only reported recently. Evolvability Search quantifies evolvability by calculating the behavioural diversity of offspring. This is achieved by sampling and evaluating the offspring of every individual in the population, only to discard all these evaluations after evolvability is calculated.
Discarding all these evaluations makes this technique extremely computationally expensive.

Another similarly expensive technique is ES-MAML \citep{song2019maml}, the evolutionary version of the meta learning technique MAML. ES-MAML selects for the adaptiveness aspect of evolvability. The adaptiveness can be calculated by applying various adaptation operators. For example, the adaptation operator can be another ES update, which includes evaluating and subsequently discarding the offspring of each individual in the population.

Finally, there are two algorithms that directly select for evolvability without a large computational cost; both are variants of the Evolution Strategy (ES) algorithm \citep{salimans2017evolution}. Evolvability ES (E-ES) \citep{gajewski2019evolvability} selects for the behavioural diversity aspect of evolvability, while Novelty Search ES (NS-ES) \citep{conti2017improving} selects for the innovation aspect of evolvability. We discuss how ES can achieve this without a large computational cost in Methodology Section.


\section{Quality Evolvability}

\begin{table*}[]
\centering
\renewcommand{\arraystretch}{1.5} 
\caption{Comparison of Quality Diversity and Quality Evolvability}
\label{tab:QDQE}
\begin{tabular}{@{}lll@{}}
\toprule
    Metric 
    & Quality Diversity (QD) 
    & Quality Evolvability (QE) \\ 
\midrule
    \makecell[l]{Approach objective\\ $\ $} 
    & \makecell[l]{Find an archive of diverse  and\\ well performing individuals} 
    & \makecell[l]{Find an individual with  diverse and\\well performing offspring} 
    \\
    \makecell[l]{Solving deceptive problems\\ $\ $} 
    & \makecell[l]{Yes, ever increasing\\ pressure to escape} 
    & \makecell[l]{Yes, constant\\ pressure to escape} 
    \\
    \makecell[l]{Illuminating search spaces\\ $\ $} 
    & \makecell[l]{Yes, will explore many ways\\to solve the problem}               
    & \makecell[l]{No, only explores a small volume\\of the search space}           
    \\
    \makecell[l]{Encouraging adaptability\\ $\ $}   
    & \makecell[l]{No, QD is allowed to use an archive of\\genetically distant individuals} 
    & \makecell[l]{Yes, QE is forced to adapt a single\\individual to many different behaviors} 
    \\ 
\bottomrule
\end{tabular}

\end{table*}

We propose Quality Evolvability (QE), an approach that aims to find individuals with both a diverse and well performing distribution of offspring. 

The main motivation for Quality Evolvability is to benefit from an increased ability to evolve in cases where looking for evolvability alone is not sufficient to find solutions to a task. This is similar to the motivation behind Quality Diversity (QD) \citep{pugh2016quality}: to benefit from the divergent, stepping stone finder nature of Novelty Search \citep{lehman2011novelty}, even in cases when seeking novelty alone is not sufficient to find solutions to a task. In short Quality Evolvability is to Evolvability Search what Quality Diversity is to Novelty Search. Even though Quality Evolvability and Quality Diversity have similar motivations, there are important differences between the two approaches, which are summarized in Table \ref{tab:QDQE}.

The unique property of Quality Evolvability is that it is forced to become more evolvable, because it needs to adapt a single individual to diverse behaviours. While Quality Diversity can benefit from developing evolvability, it is not forced to do so; it can achieve diversity by collecting a set of genetically distant individuals, without any enhanced ability to adapt.

Quality Diversity will explore a large section of the search space, aiming to find every possible behaviour that exists. For this reason, it can be used to illuminate search spaces, or to find many possible solutions to a problem. Quality Evolvability does not have this property, it focuses its search on a small volume of the search space.

Quality Diversity excels at solving deceptive problems because it will increase the pressure to find novelty until the pressure is high enough to overcome the deceptiveness. Quality Evolvability should also be helpful at escaping deceptive local minima to some degree since it aims to find offspring with diverse behaviours. However, for Quality Evolvability, this pressure is constant and not ever increasing. If the trap is large enough to allow a large amount of diversity, Quality Evolvability is expected to stay trapped.


\section{Methodology}
\label{sec:QEES}
In this section, we describe our proposed method, Quality Evolvability ES (QE-ES), which simultaneously selects for both evolvability and fitness. We also discuss the methods which QE-ES is based on, Evolvability ES (E-ES) and ES. Here, we use the term ES to refer to the recent algorithm defined in \citep{salimans2017evolution}. This is not to be confused with what has traditionally been referred to as Evolution Strategies (ES) (for example, a $1+1$ ES).

\subsection{ES}
ES is special case of the Natural Evolution Strategies (NES)\citep{wierstra2008natural} algorithm, which aims to calculate the gradient of the expected fitness of a parameterized search distribution $p_\phi$ with respect to the distribution parameters $\phi$. In case of ES, this is an isotropic normal distribution, parametarized by a center individual: $\theta = \mathcal{N}(\phi,\sigma) $ where $\sigma\in\mathbb{R}$. The cost function is defined as an expectation over the distribution $J=\mathbb{E}_{\theta\sim p(\phi)}F(\theta)$. The update rule is derived by applying the ``log-likelihood trick'' \citep{wierstra2008natural} so the gradient of the expectation can be expressed with the expectation of a gradient (eq.\ref{eq:logprob}), which can be approximated with samples.
\begin{eqnarray} \label{eq:logprob}
\nabla_\phi  \mathbb{E}_{\theta\sim p_\phi}F(\theta) = \mathbb{E}_{\theta\sim p_\phi} \{F(\theta) \nabla_\phi log p_\phi(\theta)\}
\end{eqnarray}
Previous work found that using a rank based futness shaping tend to improve performance by reducing the effect of outliers in the population \citep{salimans2017evolution}. We included this step in the description of the algorithms with the function fitness\_shaping.

\begin{algorithm}[]
\SetAlgoLined
 \KwIn{Noise standard deviation $\sigma$, initial policy parameters $\theta_0$, population size $n$, gradient optimizer, fitness function $F$}
 \For{t = 0, 1, 2, . . . } 
 {
    Sample $\epsilon_1 . . . \epsilon_n  \sim \mathcal{N}(0,I)$ \\
    $F_i=F(\theta_t +\sigma \epsilon_i)$ for $i = 1, . . . , n$  \\
    $r$ = fitness\_shaping($F$) \\
    $grad=\frac{1}{n\sigma} \sum_{i=1}^{n} r_i \epsilon_i$ \\
    $\theta_{t+1}=optimizer(\theta_t,grad)$ \\
}
\caption{ES}
\label{alg:es}
\end{algorithm}

ES has several differences compared to traditional finite difference approximators. The effect of optimizing for the expectation over a normal distribution can be imagined as using normal gradient descent on a blurred version of the fitness landscape \citep{salimans2017evolution}. Another consequence of optimizing for expected fitness instead of the fitness, is that ES aims to find solutions that are robust to perturbations \citep{lehman2018more}.

\subsection{Evolvability ES}
Evolvability ES \citep{gajewski2019evolvability} is a variant of ES which directly selects for evolvability. Evolvability is defined as either the variance or the entropy of the distribution of behaviour (as measured by the behaviour characterization (BC) function). In this work, we use the variance, because it is simpler and seems to work equally as well as entropy \citep{gajewski2019evolvability}. Therefore, evolvability is defined as: 
\begin{eqnarray} \label{eq:evolvability}
Evolvability(\phi) = \mathbb{E}_{\theta\sim p_\phi}(BC(\theta)-BC_{mean})^2
\end{eqnarray}
The simplest way to create an ES variant that directly selects for evolvability is to replace fitness with evolvability in the cost function: $J=\mathbb{E}_{\theta\sim p(\phi)}Evolvability(\theta)$. However, this way requires the evaluation of the evolvability of the whole population, which requires sampling the offspring of each individual in the population, making this approach computationally expensive.

\cite{gajewski2019evolvability} recognized that there is a different way of creating an ES algorithm that directly selects for evolvability. Evolvability is itself defined as an expectation (both the variance or entropy versions), so the log-likelihood trick which is used to derive the ES update rule can be applied directly to evolvability, instead of the expected evolvability. The resulting algorithm, Evolvability ES, has the cost function: $J=Evolvability(\phi)$. Evolvability ES no longer needs to calculate the evolvability of every individual, making Evolvability ES as fast as normal fitness based ES. It is important to note that the new algorithm no longer has the blurring and robustness seeking property of ES, and is more like a traditional finite difference approximator.

\begin{algorithm}[]
\SetAlgoLined
\KwIn{Noise standard deviation $\sigma$, initial policy parameters $\theta_0$, population size $n$, gradient optimizer, behaviour characterization $BC$ }
 \For{t = 0, 1, 2, . . . } 
 {
    Sample $\epsilon_1 . . . \epsilon_n  \sim \mathcal{N}(0,I)$ \\
    ${EVO}_i = (BC(\theta_t + \sigma \epsilon_i)-BC_{mean})^2$ for $i = 1, . . . , n$  \\
    $r$ = fitness\_shaping($EVO$) \\
    $grad=\frac{1}{n\sigma} \sum_{i=1}^{n} r_i \epsilon_i$ \\
    $\theta_{t+1}=optimizer(\theta_t,grad)$ \\
}
\caption{Evolvability ES}
\label{alg:EvoES}
\end{algorithm}

\subsection{Quality Evolvability ES}

Our proposed method Quality Evolvability ES (QE-ES) combines the objectives of ES and E-ES, to simultaneously optimize for both evolvability and fitness. To achieve this we use non-dominated sorting (nd\_sort) on the evolvability and fitness objectives. Non-dominated sorting is done the same way as in the well-known NSGA-II \citep{deb2002fast} algorithm. Individuals are first sorted by which non-dominated front they belong to, then a crowding metric is used to sort the individual within the fronts. The crowding metric ensures that a diverse set of evolvability-fitness trade-offs are maintained. 

\begin{algorithm}[]
\SetAlgoLined
\KwIn{Noise standard deviation $\sigma$, initial policy parameters $\theta_0$, population size $n$, gradient optimizer, fitness function $F$, behaviour characterization $BC$}
 \For{t = 0, 1, 2, . . . } 
 {
    Sample $\epsilon_1 . . . \epsilon_n  \sim \mathcal{N}(0,I)$ \\
    $F_i=F(\theta_t +\sigma \epsilon_i)$ for $i = 1, . . . , n$  \\
    ${EVO}_i = (BC(\theta_t + \sigma \epsilon_i)-BC_{mean})^2$ for $i = 1, . . . , n$  \\
    $r$ = fitness\_shaping(nd\_sort($F,EVO$)) \\
    $grad=\frac{1}{n\sigma} \sum_{i=1}^{n} r_i \epsilon_i$ \\
    $\theta_{t+1}=optimizer(\theta_t,grad)$ \\
}
\caption{Quality Evolvability ES}
\label{alg:QES}
\end{algorithm}

\section{Experiments}

\begin{figure}

    \centering     

    \begin{subfigure}[Deceptive Ant task]
    {
    \includegraphics[width=38mm]{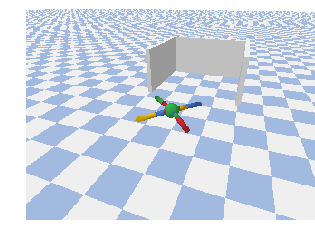}
    }
    \end{subfigure}
    \begin{subfigure}[Deceptive Humanoid task]
    {
    \includegraphics[width=38mm]{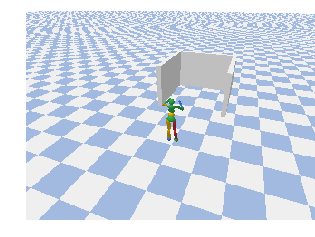}
    }
    \end{subfigure}
    \caption{For the deceptive variant of the tasks, a trap box is put in front of the agent. This obstacle creates a deceptive local optimum. Once the agent discovers how to walk into the box, it cannot improve further without developing the ability to walk around the obstacle. The sides of the trap however make walking around it difficult, requiring the fitness to decrease first. Greedy objective based algorithms are susceptible to such deceptive local optima.}
    \label{fig:deceptive_images}
\end{figure}

We evaluated our method on the robotics locomotion tasks Ant and Humanoid. An evaluation is comprised of running a full episode until the maximum number of allowed time steps is reached or until the robot falls over. In the default version of the task, fitness is defined as the distance traveled in the $x$ direction, while behaviour is characterized by the final $x$ and $y$ coordinates of the robot.

We used the more accessible PyBullet implementation of the environments, which has a free software license, rather than the commonly used MuJuCo commercial implementation. An important difference between the implementations is that PyBullet implements the observations differently in the case of the Humanoid environment, providing only a $\in\mathbb{R}^{44}$ observation vector compared to MuJuCo which provides $\in\mathbb{R}^{376}$ dimensional observation, containing a more detailed motion state of the robot. This makes the Humanoid experiments more challenging in PyBullet. While with previous experiments using the MuJuCo version of Humanoid, ES was able to find policies that reliably solved the task. In our experiments, ES only finds policies that sometimes solved the task, but not reliably, and not every training run finds such policies. With the Ant environment, there are no significant differences between the implementations, and we could replicate previous results.

Evolution evolves the parameters of a 2-hidden-layer fully connected neural network with 256 neurons for both hidden layers, resulting in a total of 75k parameters for the Ant environment and 81k for the Humanoid environments. The network maps the observations to the actions, with both being real numbers.

For all experiments, we used the same hyperparameter values used in \citep{gajewski2019evolvability}. The population size was 10,000, the noise standard deviation $\sigma$ was 0.02. We used the Adam optimizer with a learning rate $\alpha$ of 0.01 and L2 regularization of 0.005.
We used mirrored sampling and centered rank normalization. The Ant experiments were run for 200 generations, the more difficult Humanoid experiments were run for 800 generations.

ES methods are less sample efficient compared to reinforcement learning methods. In our experiments, a single run consisted of simulating several billion steps. For example, in the case of the successful humanoid run, we have a population size of 10,000, which is evolved for 800 generations, where each evaluation takes around 800 time steps. However, ES parallelizes well (linear scaling \citep{salimans2017evolution}). We run our experiments on a distributed CPU cluster using 120 cores for each run. With this setting, a Humanoid run took around 1-2 days, depending on the average episode length, while an Ant run took around 6-12 hours. We used the base ES implementation provided by \citep{gajewski2019evolvability}.


To test different properties of our method, we used three modified versions of the environments.

\subsubsection{Task 1: Normal Locomotion}
For the first experiment, we used the default environments, which we simply refer to as Ant and Humanoid. In this task, evolvability and fitness are aligned. Purely maximizing evolvability will also result in high fitness. This is because evolvability is measured as the variation of the final positions, and the way to maximize evolvability is to learn to walk far in every direction. This includes the forward direction, which means that maximizing evolvability will also maximize fitness.

\subsubsection{Task 2: Directional Locomotion}
For the second experiment, we used modified environments, which we call Directional Ant and Directional Humanoid. Our goal with this experiment is to test our method in a case where evolvability and fitness are unaligned. For these experiments, for each episode, we randomly select a direction from 8 possible directions that are evenly distributed around a circle in 45 degree increments for each episode. The networks receive this direction, represented as a unit vector, as an observation which increases the number of observations by 2. 

To perform well in this task, the agent needs to turn in the correct direction and walk. Maximizing diversity alone is not enough to achieve high fitness anymore; the agent also needs to learn to walk in the correct direction. Evolvability ES is expected to have zero mean fitness on this task because it equally prefers every direction. 

\subsubsection{Task 3: Deceptive Locomotion}
For the third experiment, we used a deceptive variant of the environments, similarly to previous work \citep{conti2017improving}. We call these environments Deceptive Ant and Deceptive Humanoid. A U-shaped trap is placed in front of the agent, which allows it to make progress for a while until it runs into a wall (Fig.\ref{fig:deceptive_images}). The walls on the side prevent the agent to easily go around the obstacle, forming a trap, a local optimum, which is hard to escape. For Deceptive Humanoid, we used the same trap box configuration as in \citep{conti2017improving}, for the Deceptive Ant environment we slightly increased the dimensions of the box from 3 meters to 4 meters in order to make the trap large enough for the physically wider Ant robot.

\section{Results}

We use three kinds of plots to present and compare the result achieved with the different algorithms. To show the behaviour of the population in the final generation of a single run, we use a 2d histogram which shows the frequency of the final $x$, $y$ positions of the robot. The box and strip plots show the mean of the population in the last generation (fitness or distance walked), over repeated runs. Finally, the learning curves show the mean of the population averaged over repeated runs throughout the generations. 

For each task and each algorithm, we repeated the runs at least 10 times. We found that there is a high variance in the results with different random seeds, especially in the case of the more difficult Humanoid task. This agrees with previously reported results in this domain \citep{conti2017improving}. 


\begin{figure}
    \centering 
    \begin{subfigure}[Humanoid, ES]
    {
        \includegraphics[width=38mm]{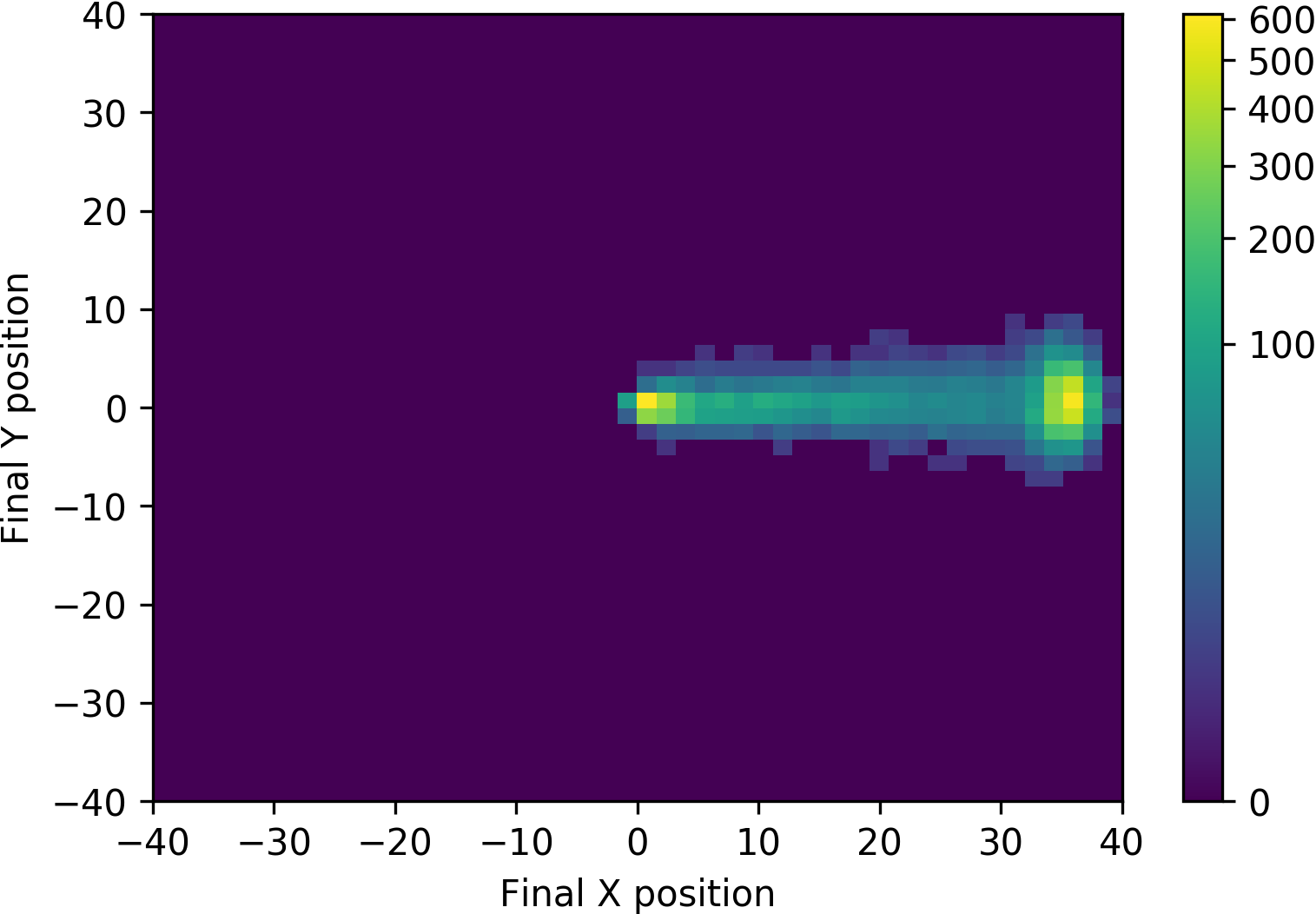}
    }
    \end{subfigure}
    \hfill
    \begin{subfigure}[Humanoid, E-ES]
    {
        \includegraphics[width=38mm]{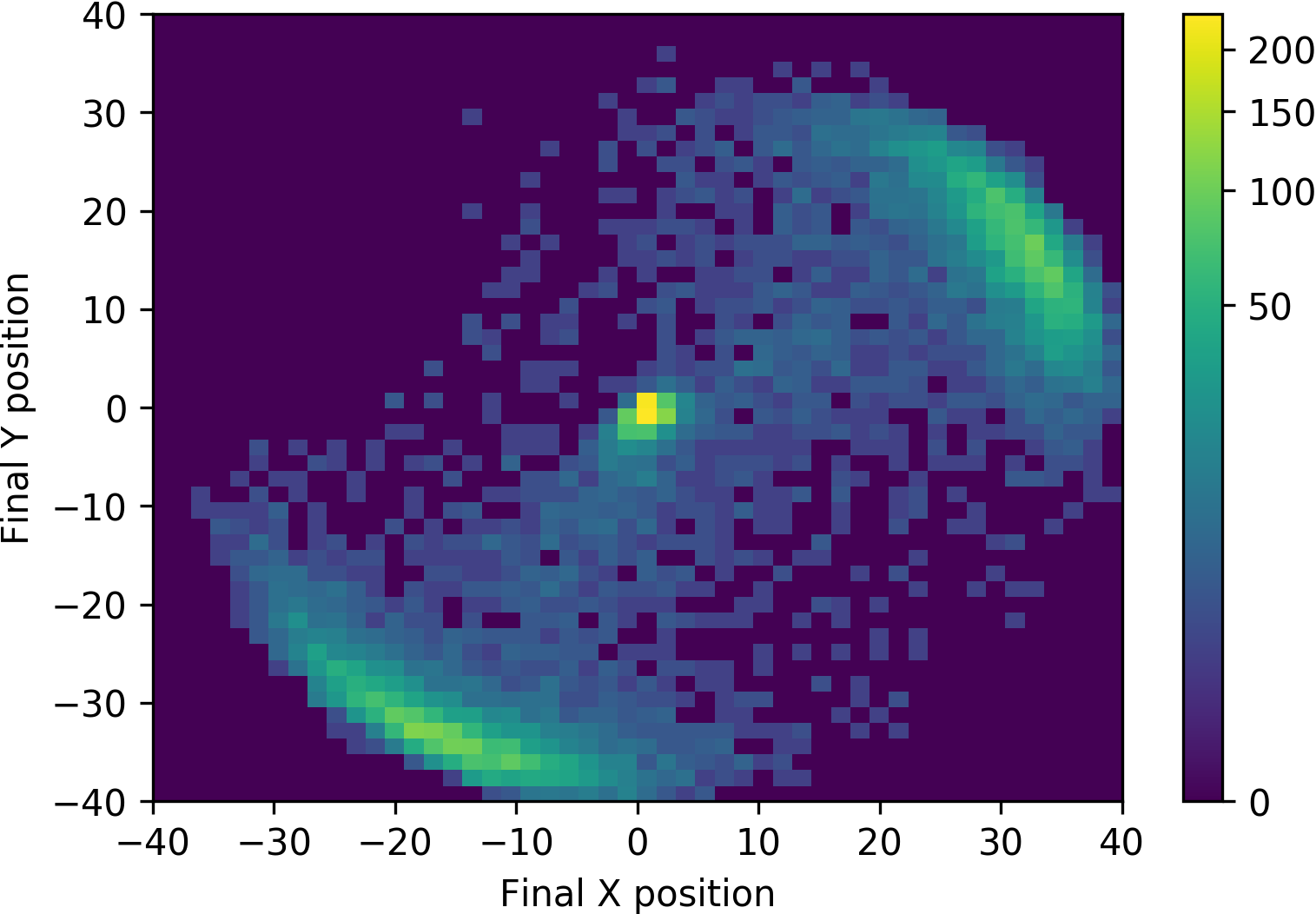}
    }
    \end{subfigure}
     \begin{subfigure}[Humanoid, QE-ES]
    {
        \includegraphics[width=38mm]{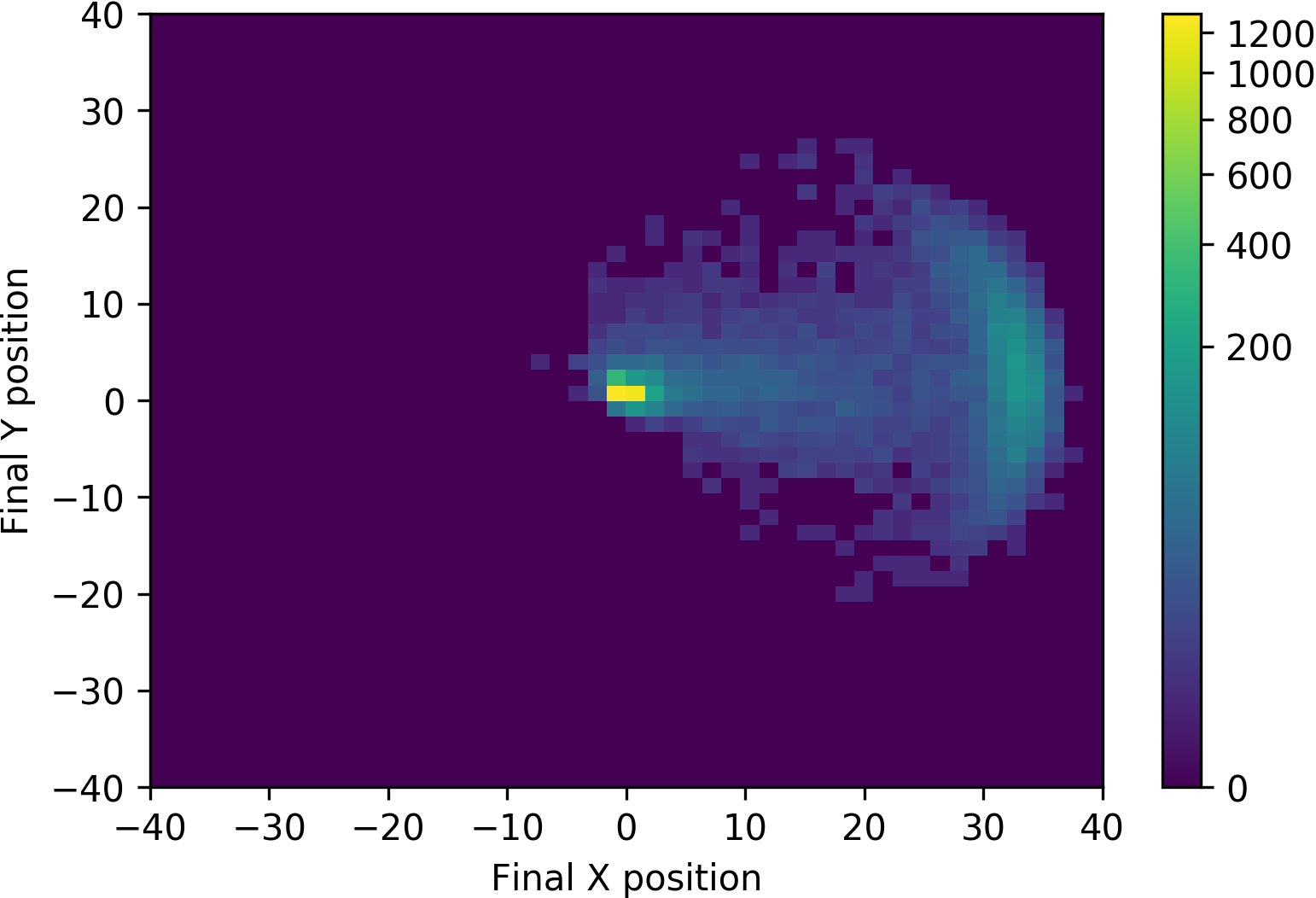}
    }
    \end{subfigure}

    \caption{2D histogram of the final positions of the population from the last generation for the Humanoid environment, which highlights the different aims of the three algorithms. (a) ES only cares about fitness, thus only finds policies that walk forward. (b) E-ES only cares about diversity and finds policies that walk in various directions. (c) QE-ES cares about both fitness and diversity, so it finds policies that walk forward in a diverse way.}
     \label{fig:exp1_histograms}
\end{figure}

\begin{figure}
    \centering 
    \begin{subfigure}[Ant]
    {
        \includegraphics[width=38mm]{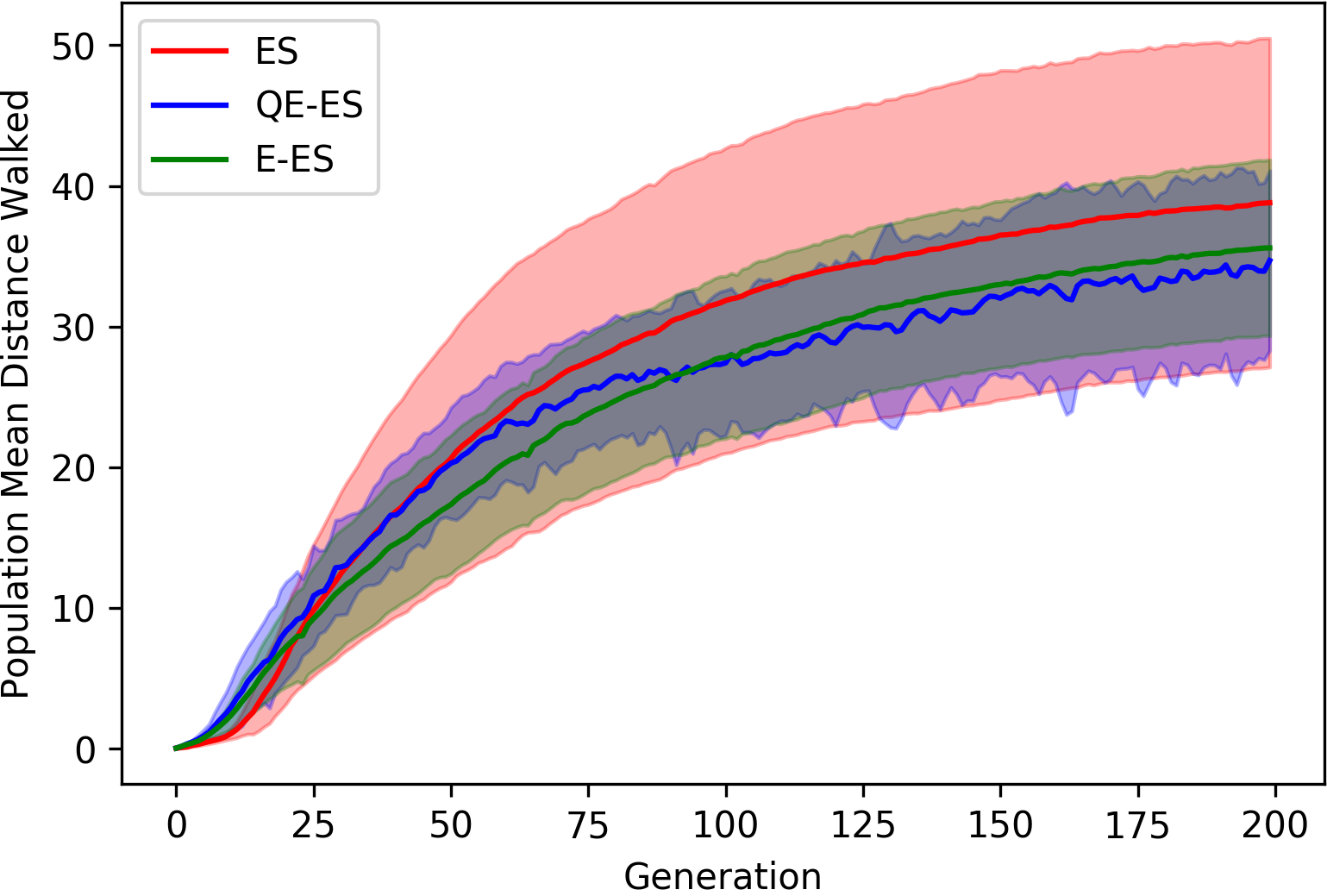}
    }
    \end{subfigure}
    \hfill
    \begin{subfigure}[Humanoid]
    {
        \includegraphics[width=38mm]{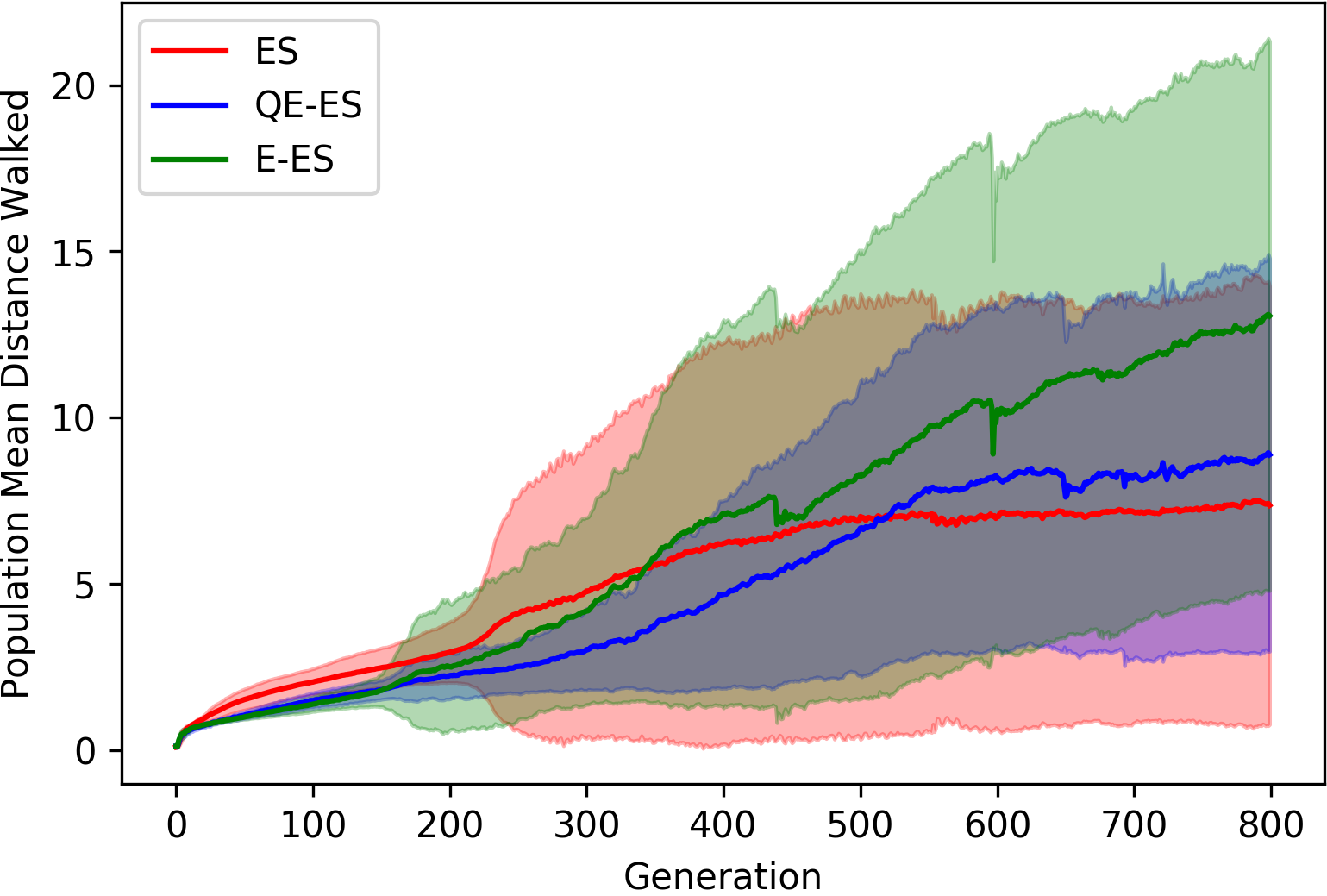}
    }
    \end{subfigure}
    \begin{subfigure}[Ant]
     {
         \includegraphics[width=38mm]{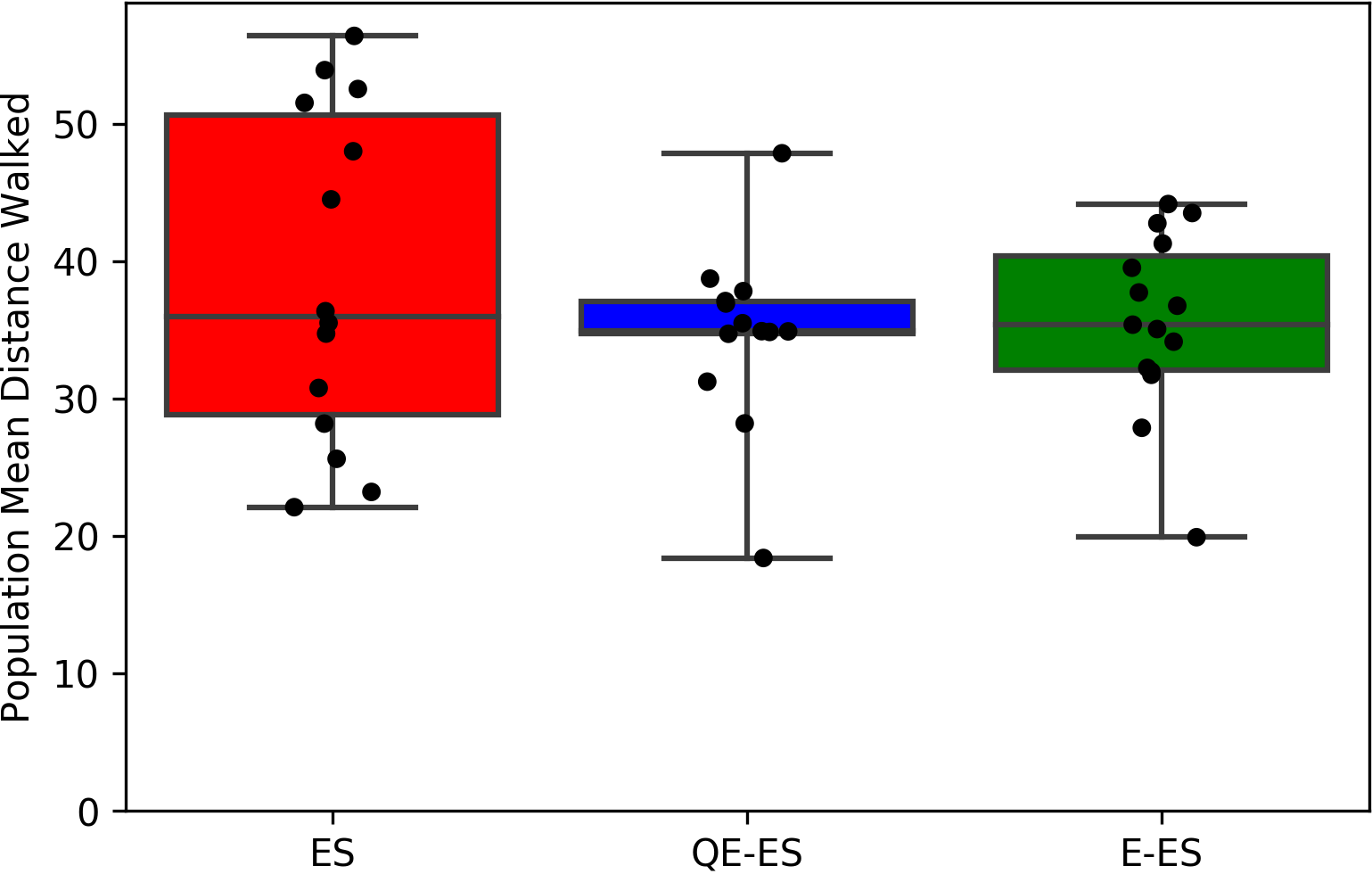}
     }
     \end{subfigure}
     \hfill
     \begin{subfigure}[Humanoid]
     {
         \includegraphics[width=38mm]{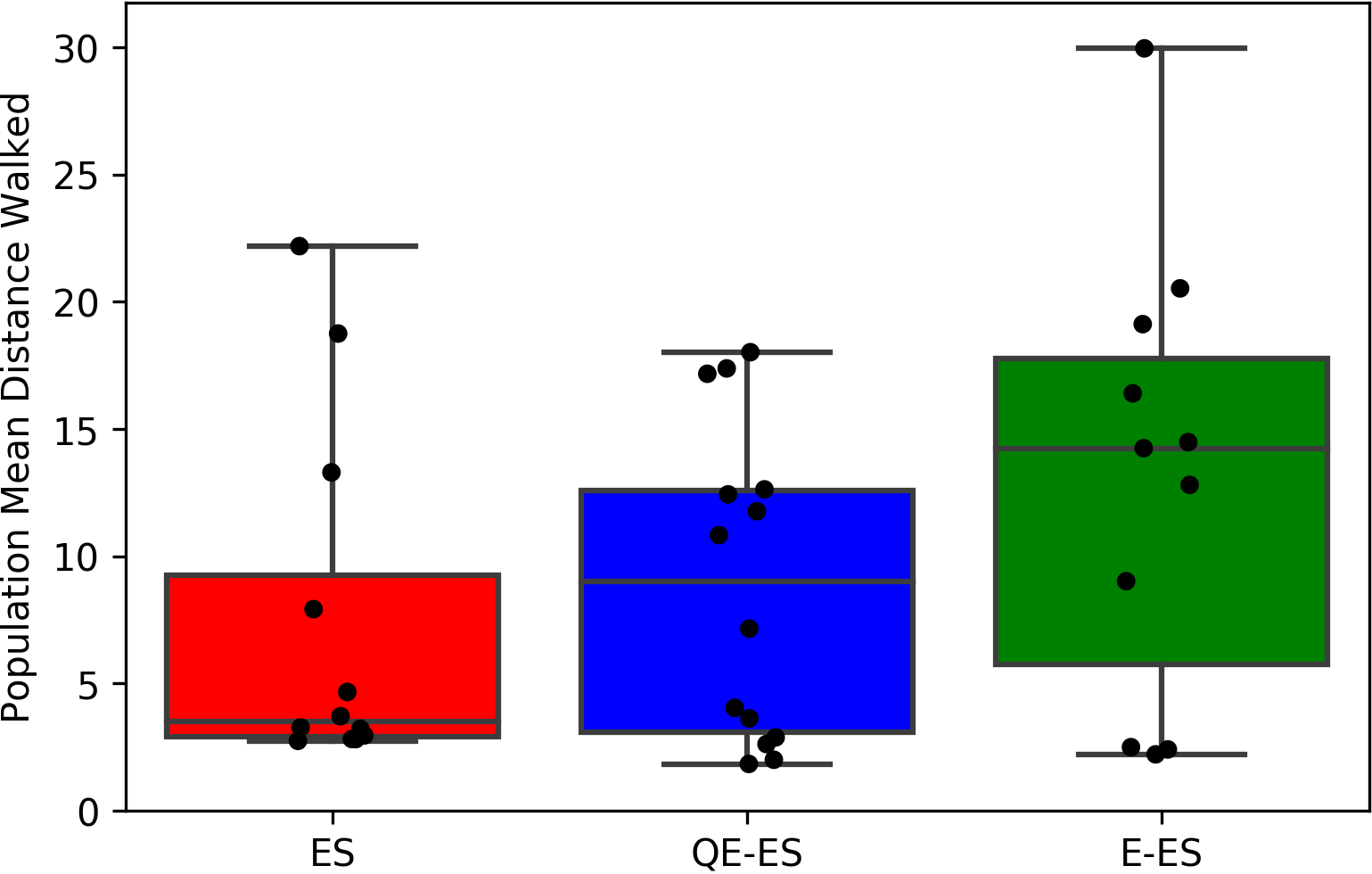}
     }
     \end{subfigure}
    \caption{Mean distance walked for the Ant (a,c) and Humanoid (b,d) tasks where evolvability and fitness are aligned. The shaded area corresponds to the standard deviation. In (a,c), E-ES finds policies that walk almost as far as ES. As evolvability and fitness are aligned in this task, QE-ES performs similarly to E-ES. In (b,d), we found that E-ES finds policies that on average walk further than ES (looking for evolvability resulted in faster learning). The performance of QE-ES is between the two since it simultaneously looks for fitness (aim of ES) and evolvability (aim of E-ES).}
      
    \label{fig:exp1_curves}
\end{figure}

\subsubsection{Task 1: Normal Locomotion}
The first experiment was done with the default environments where fitness and evolvability are aligned (Fig \ref{fig:exp1_curves}.). In case of the Ant environment we got similar results as presented in the literature previously \citep{gajewski2019evolvability}. ES performs slightly better then Evolvability ES, finding policies which on average walk further. In the humanoid environment however, Evolvability ES performs better than ES, suggesting that in some cases, searching for evolvability alone results in better task performance than searching for fitness. 
This result is similar to how novelty search can outperform objective based search. By encouraging evolvability, Evolvability ES can acquire the ability to discover skills which are not immediately useful for fitness, but can lead to progress in the future.
Since Quality Evolvability ES is a mix of the two methods, in environments where evolvability and fitness are aligned, the performance of QE-ES is expected to be between the two methods (ES and E-ES), which is what we observe. When we look at the diversity of the population, it is also as expected, E-ES having the most diversity, ES having the least, while QE-ES is in between the two (see Fig. \ref{fig:exp1_histograms}).

\begin{figure}
    \centering 
    \begin{subfigure}[Directional Ant]
    {
        \includegraphics[width=38mm]{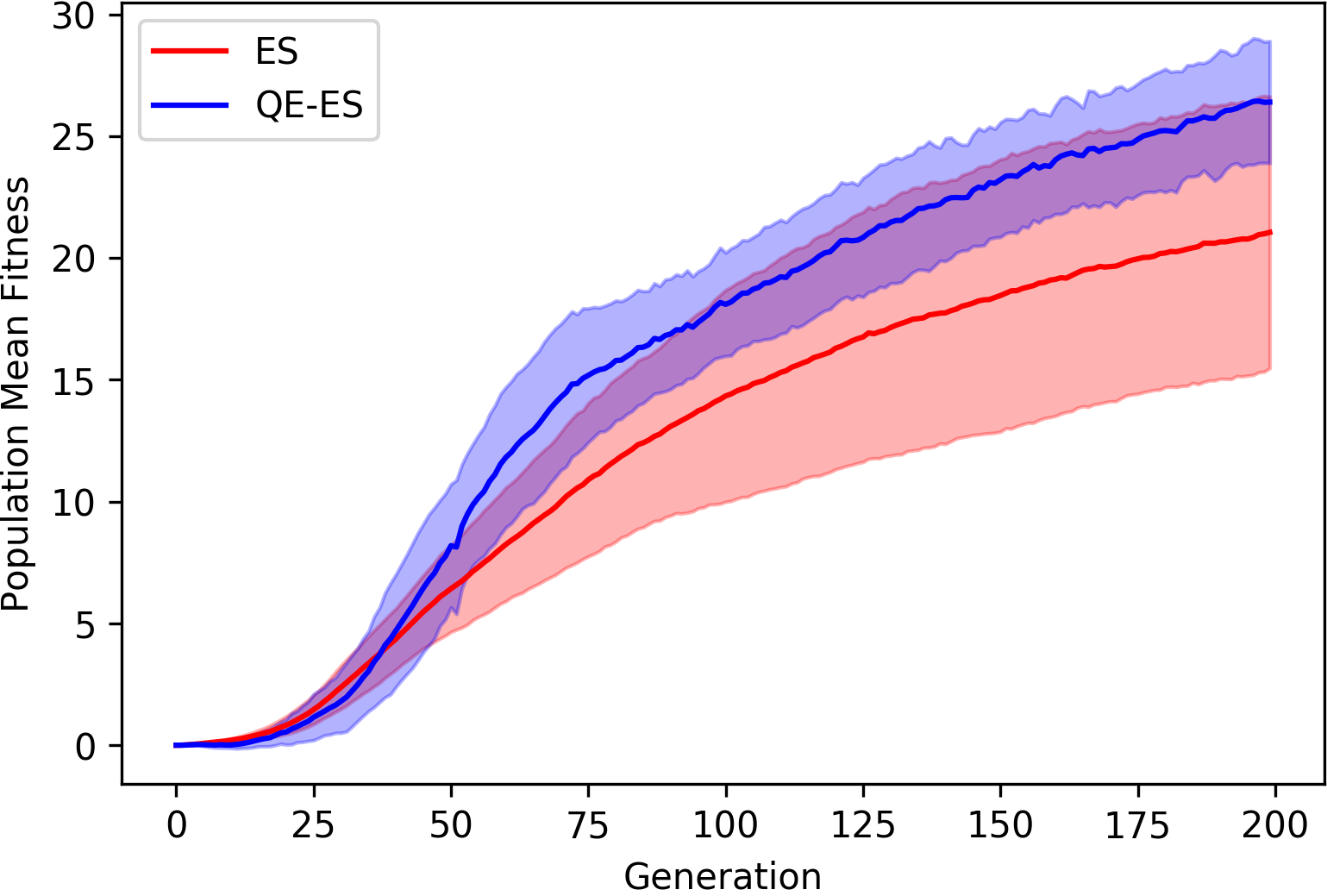}
    }
    \end{subfigure}
    \hfill
    \begin{subfigure}[Directional Humanoid]
    {
        \includegraphics[width=38mm]{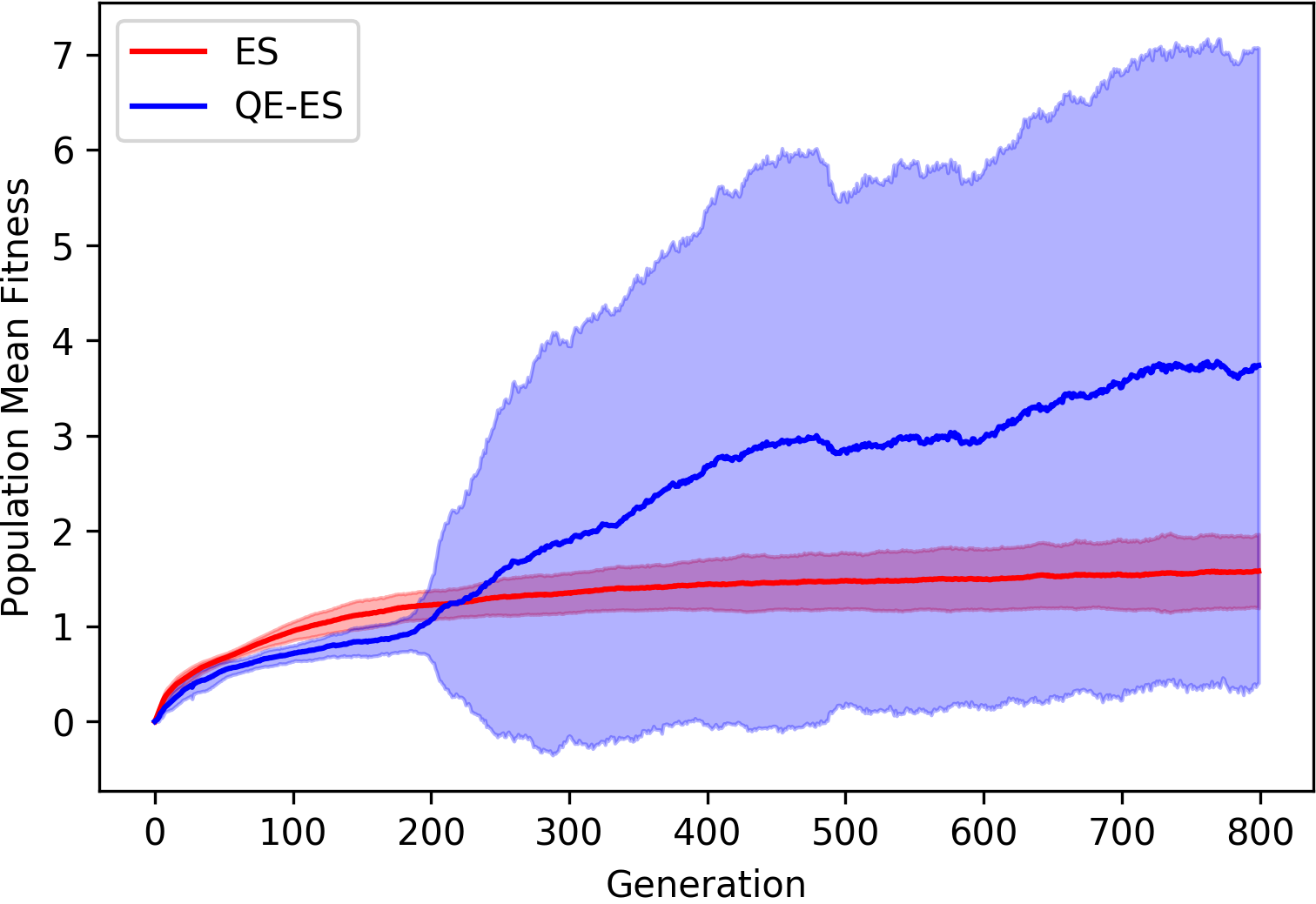}
    }
    \end{subfigure}
    \begin{subfigure}[Directional Ant]
    {
     \includegraphics[width=38mm]{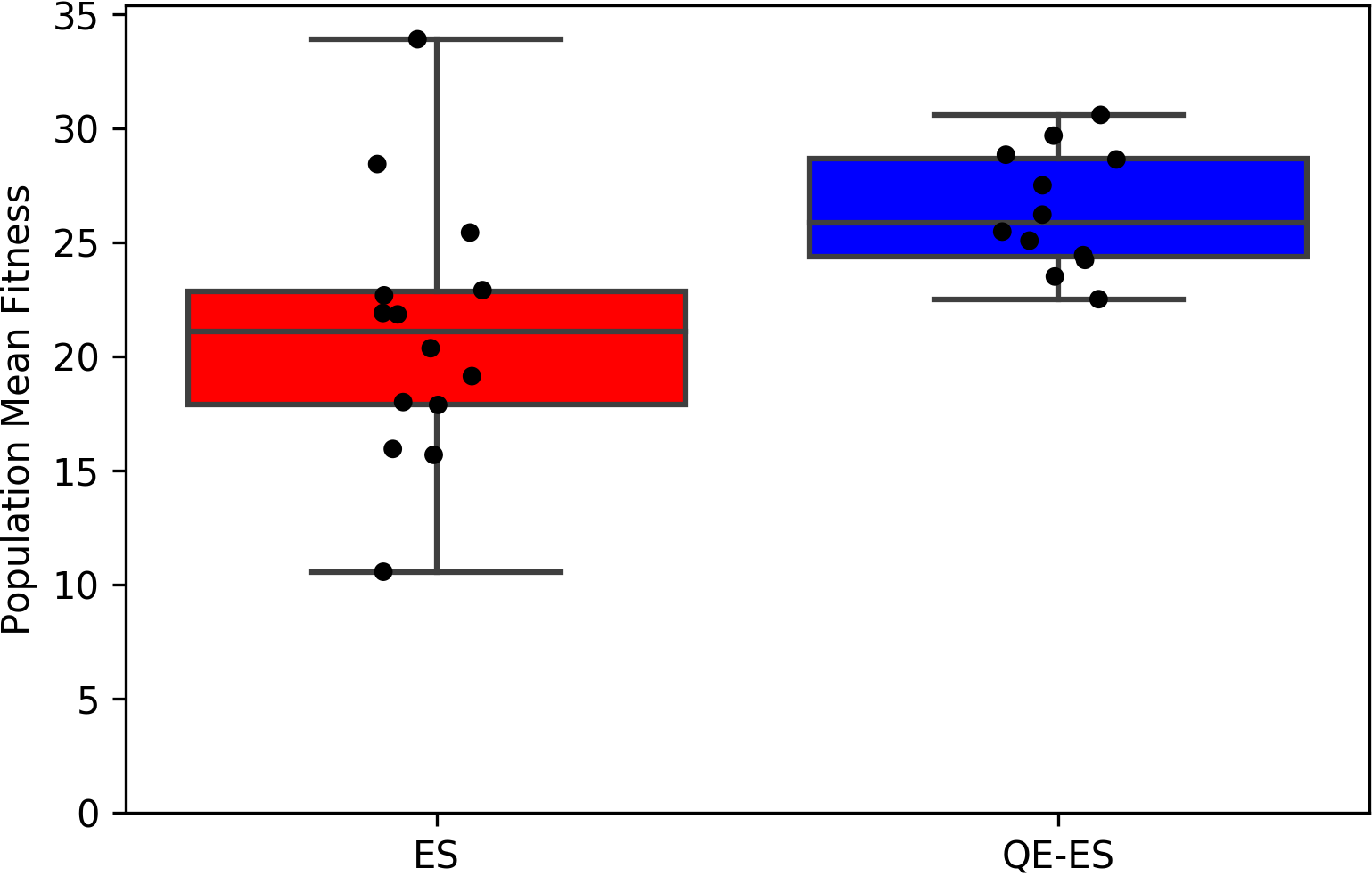}
    }
    \end{subfigure}
    \hfill
    \begin{subfigure}[Directional Humanoid]
    {
     \includegraphics[width=38mm]{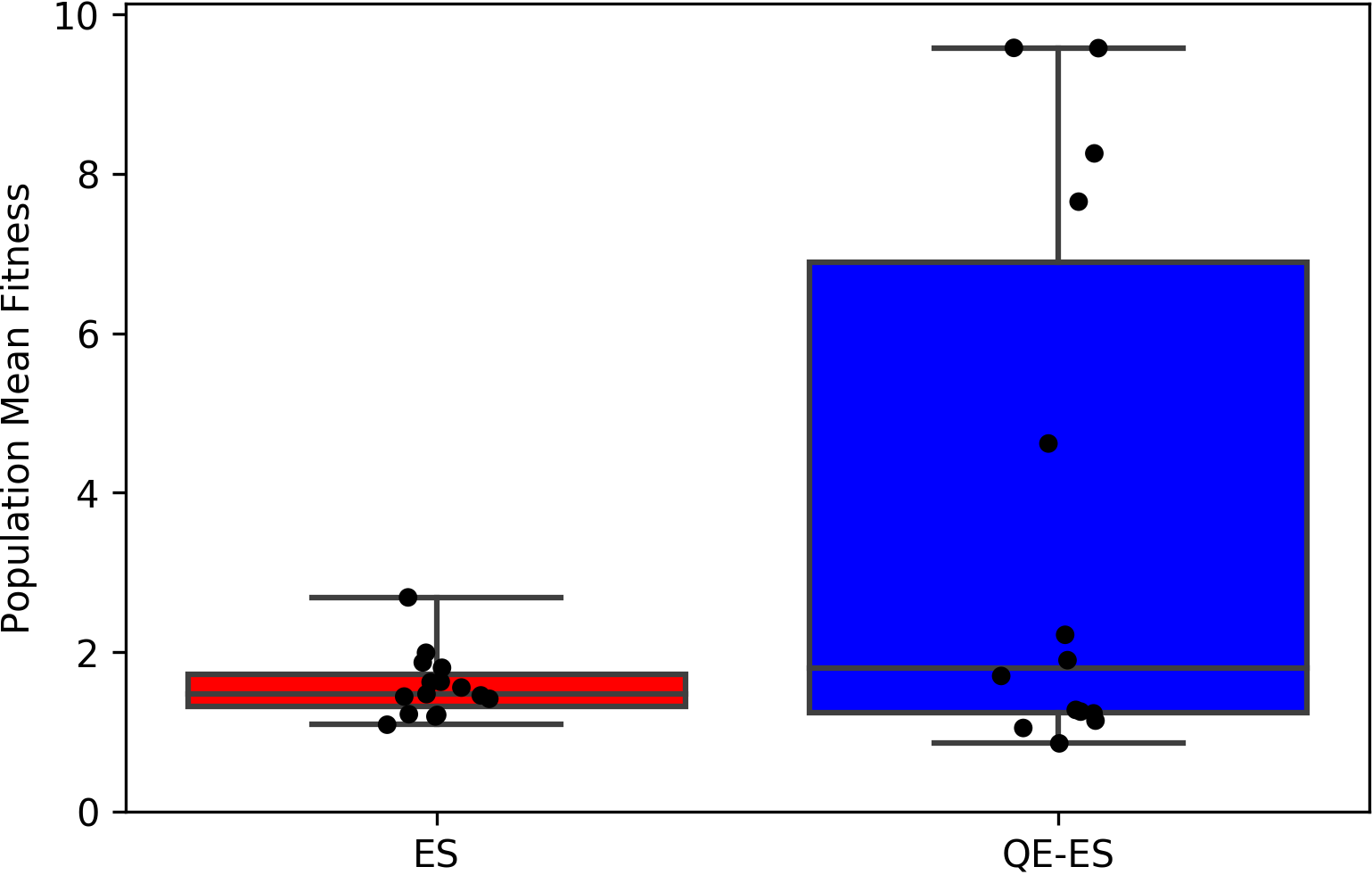}
    }
    \end{subfigure}
    \caption{Mean fitness for the Directional Ant (a,c) and Directional Humanoid (b,d) tasks, where evolvability and fitness are not aligned. In (a), both algorithms are able to find good policies, but QE-ES receives higher fitness on average. In the more difficult task (b), ES fails to learn to walk and turn in the right direction at the same time, while QE-ES is able to makes progress.}
    \label{fig:exp2_curves}
\end{figure}

\subsubsection{Task 2: Directional Locomotion}
The second experiment was done in the directional environments, where evolvability and fitness are no longer aligned (see Fig \ref{fig:exp2_curves}). Evolvability ES has zero expected fitness on this problem since it completely ignores the fitness function. Both on the Directional Ant and the Directional Humanoid tasks, QE-ES achieves higher fitness than ES. On the Directional Ant environment, the difference is relatively small and both methods can find good policies. On the more challenging Directional Humanoid environment, the challenge to simultaneously learn to walk and to turn in the correct direction proved to be too difficult for ES. In contrast, QE-ES makes some progress. These results show that selecting for evolvability can not only result in faster learning, but it can also allow making progress in cases where objective based learning gets stuck.

\begin{figure}
    \centering 
    \begin{subfigure}[Deceptive Ant]
    {
        \includegraphics[width=38mm]{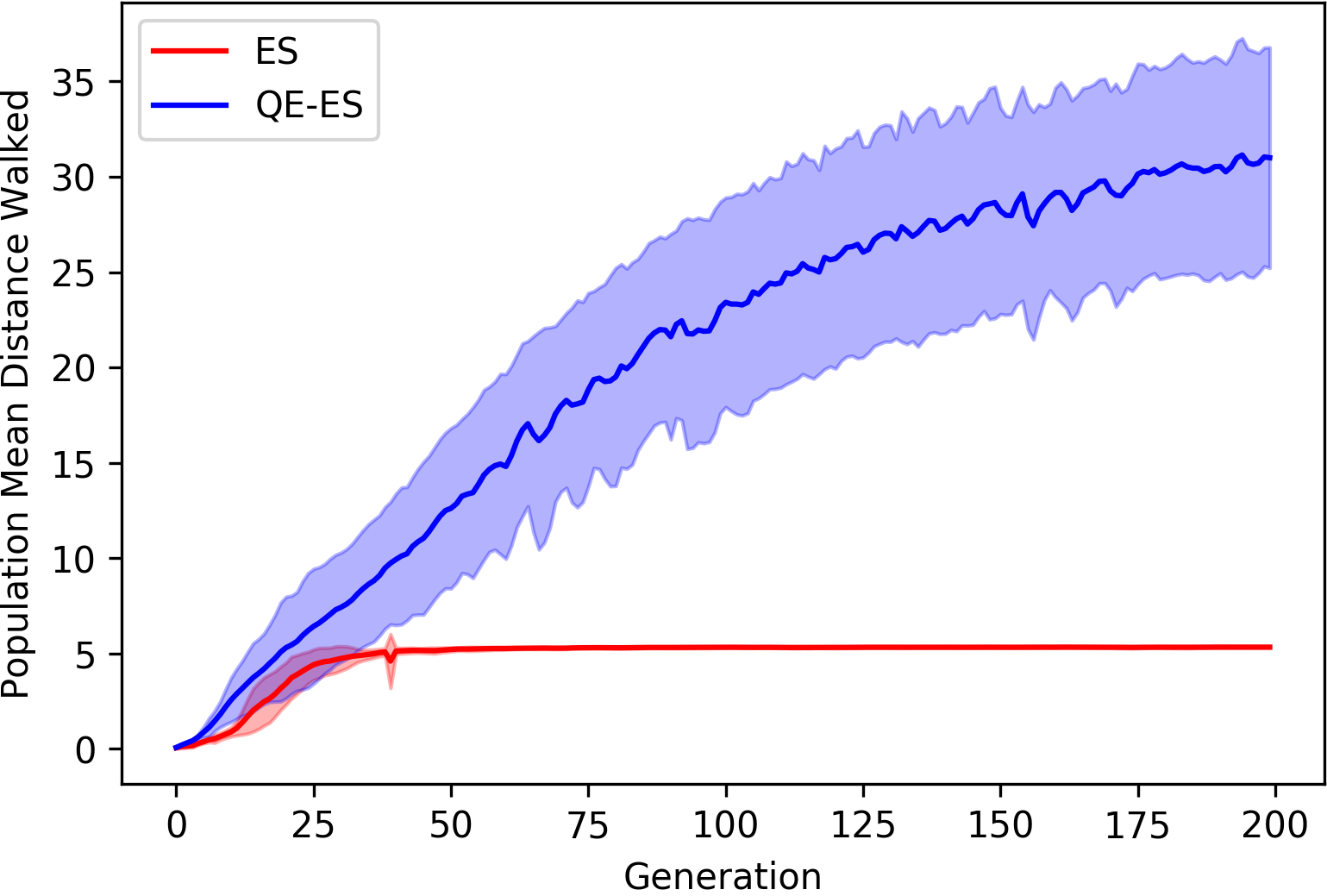}
    }
    \end{subfigure}
    \hfill
    \begin{subfigure}[Deceptive Humanoid]
    {
        \includegraphics[width=38mm]{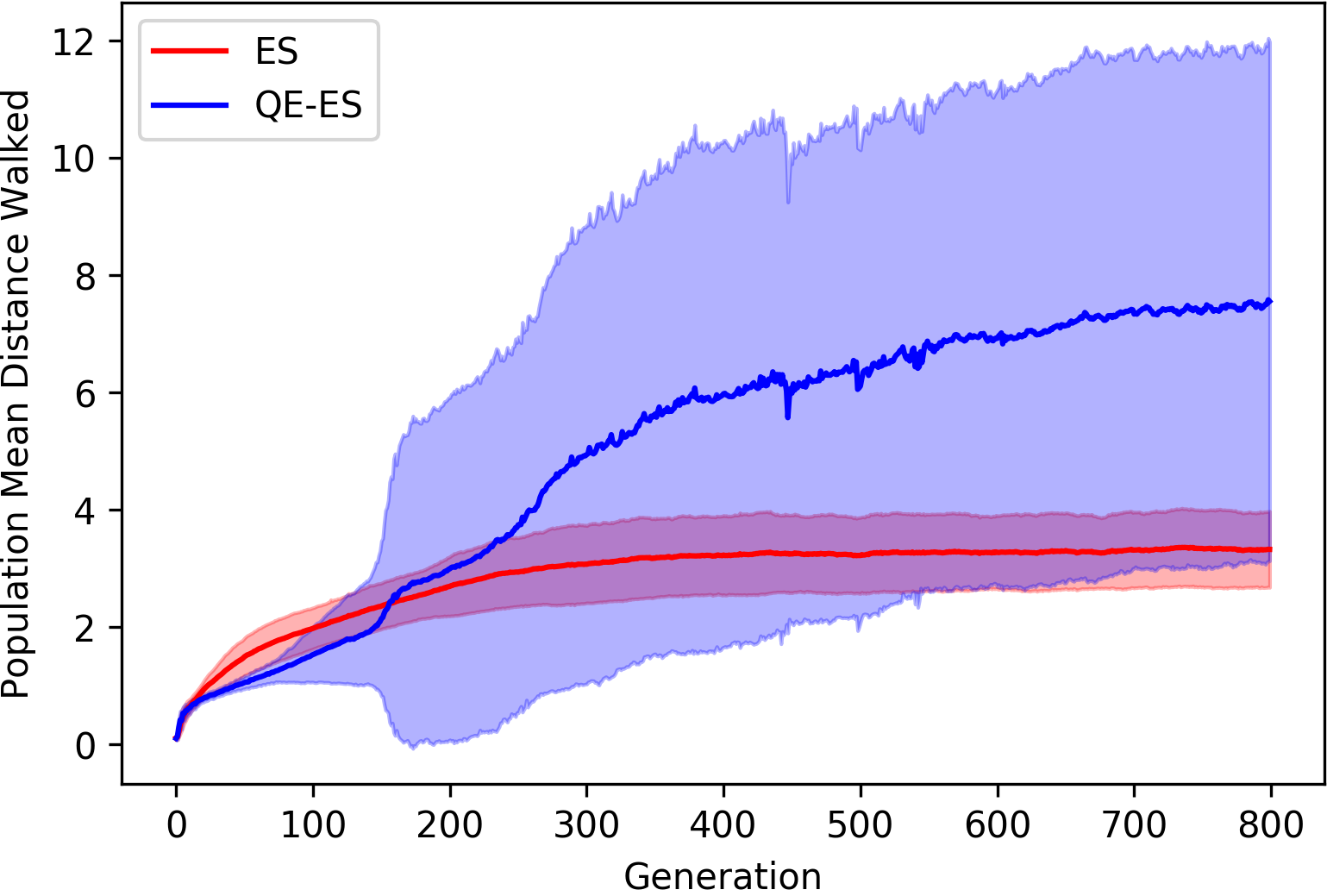}
    }
    \end{subfigure}
    \begin{subfigure}[Deceptive Ant]
    {
     \includegraphics[width=38mm]{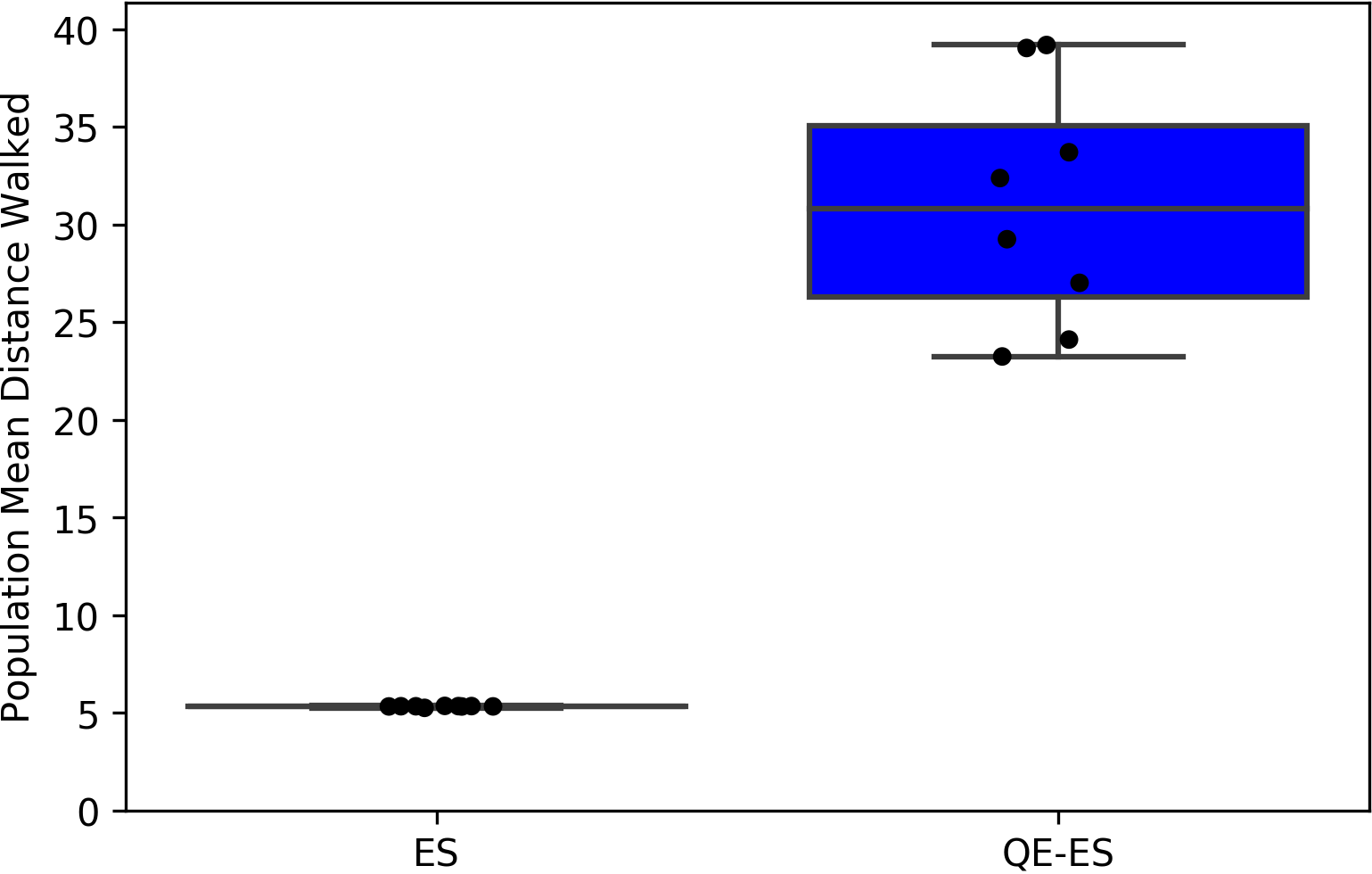}
    }
    \end{subfigure}
    \hfill
    \begin{subfigure}[Deceptive Humanoid]
    {
     \includegraphics[width=38mm]{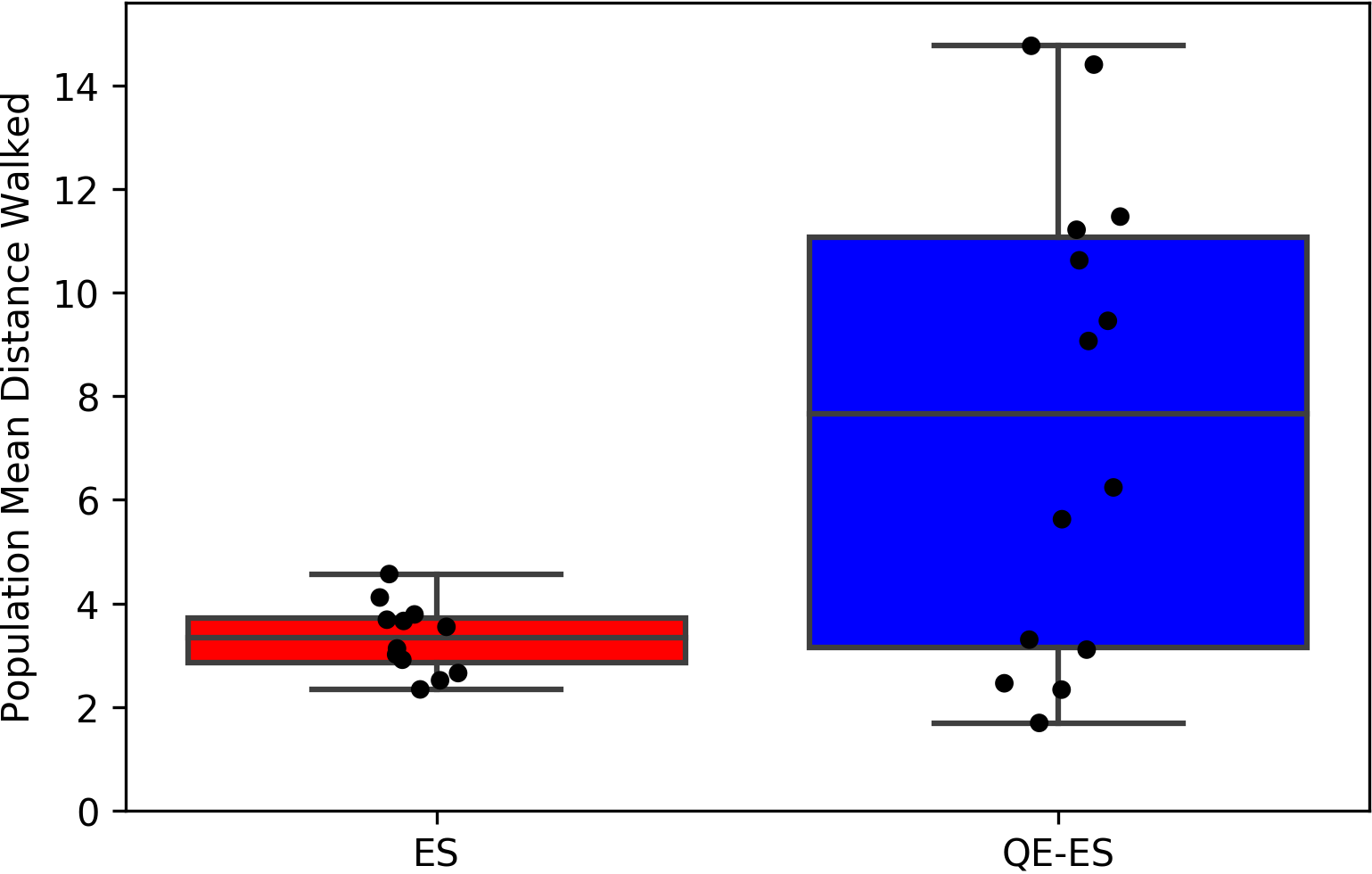}
    }
    \end{subfigure}
    \caption{Mean distance walked for the Deceptive Ant (a,c) and Deceptive Humanoid (b,d) tasks. For both environments, ES gets trapped every time. For Ant (a,c), QE-ES manages to escape the trap for every run. For the more difficult Humanoid (b,d), QE-ES escapes for the majority of the runs.}
    \label{fig:exp3_curves}
\end{figure}

\begin{figure}

    \centering 
    \begin{subfigure}[Deceptive Ant, ES]
    {
        \includegraphics[width=38mm]{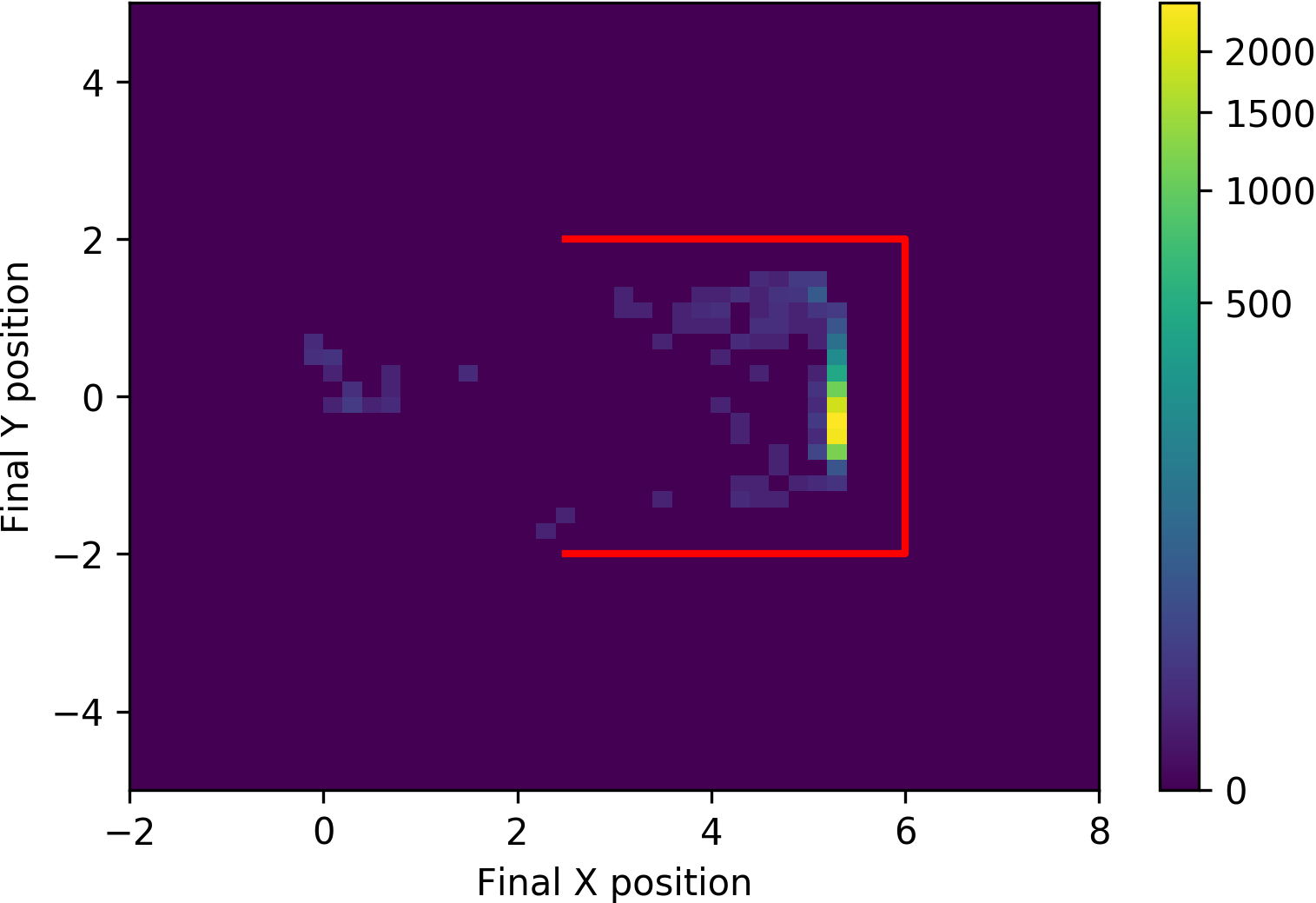}
    }
    \end{subfigure}
    \hfill
    \begin{subfigure}[Deceptive Ant, QE-ES]
    {
        \includegraphics[width=38mm]{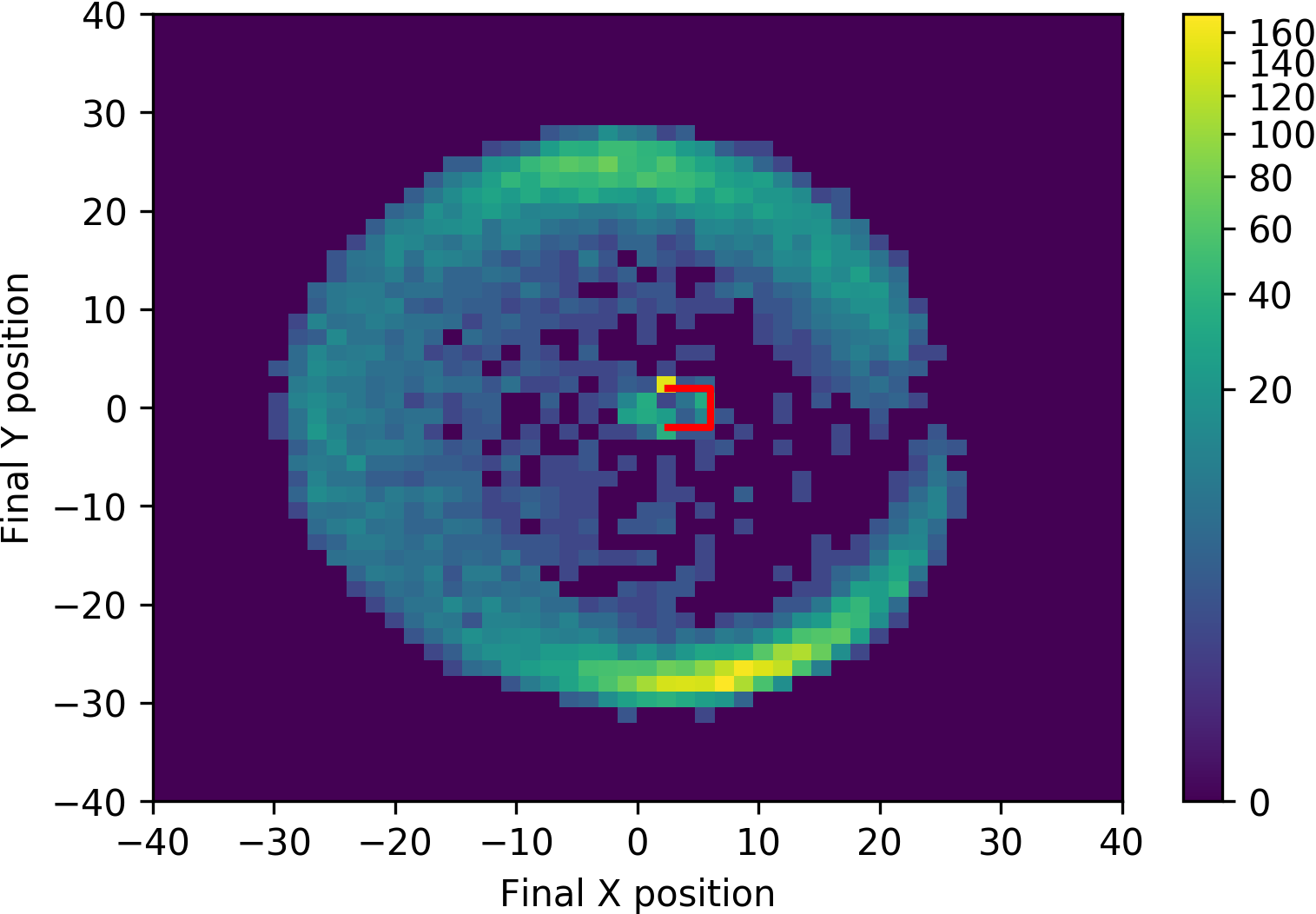}
    }
    \end{subfigure}
    \begin{subfigure}[Deceptive Humanoid, ES]
    {
        \includegraphics[width=38mm]{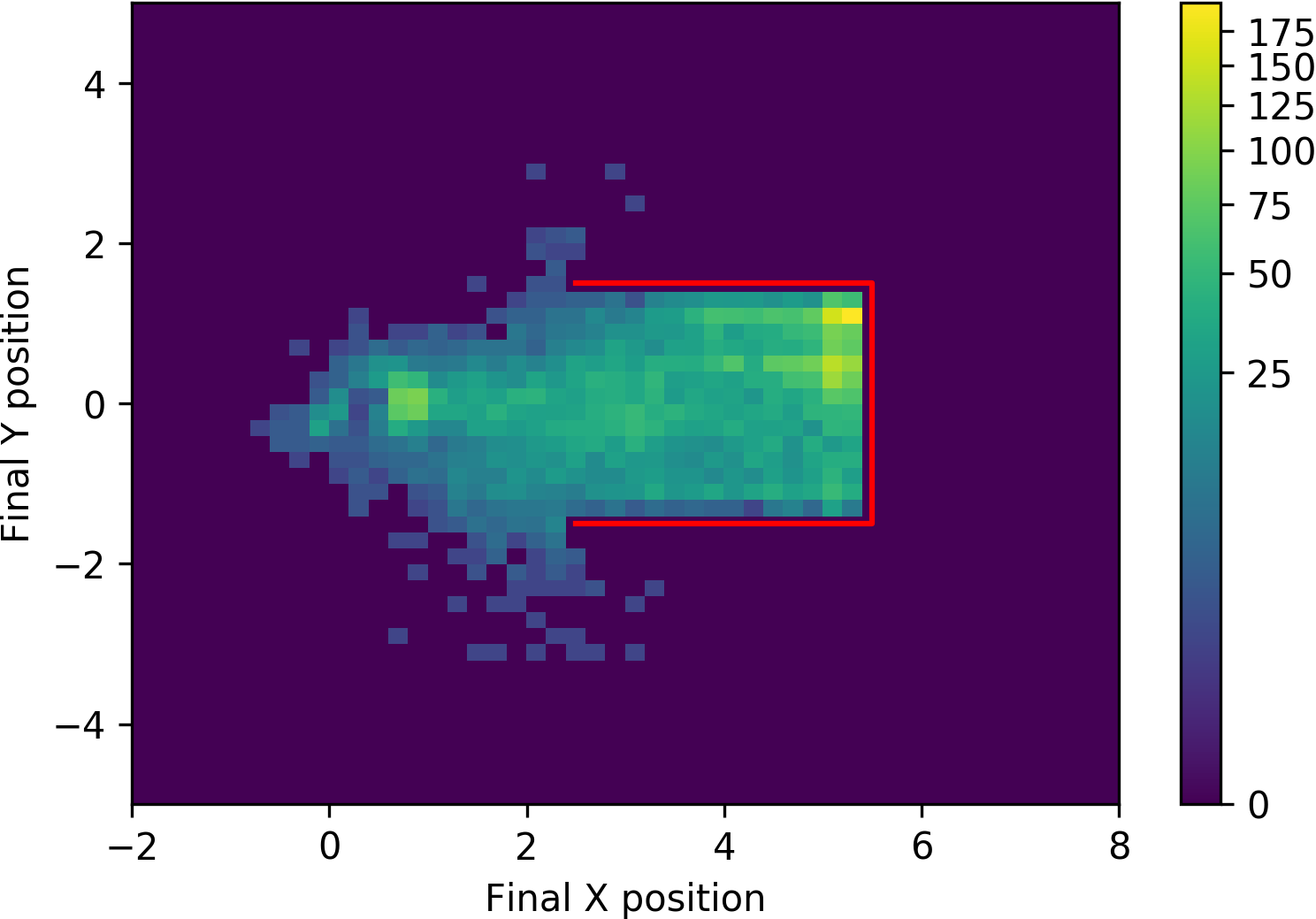}
    }
    \end{subfigure}
    \hfill
    \begin{subfigure}[Deceptive Humanoid, QE-ES]
    {
        \includegraphics[width=38mm]{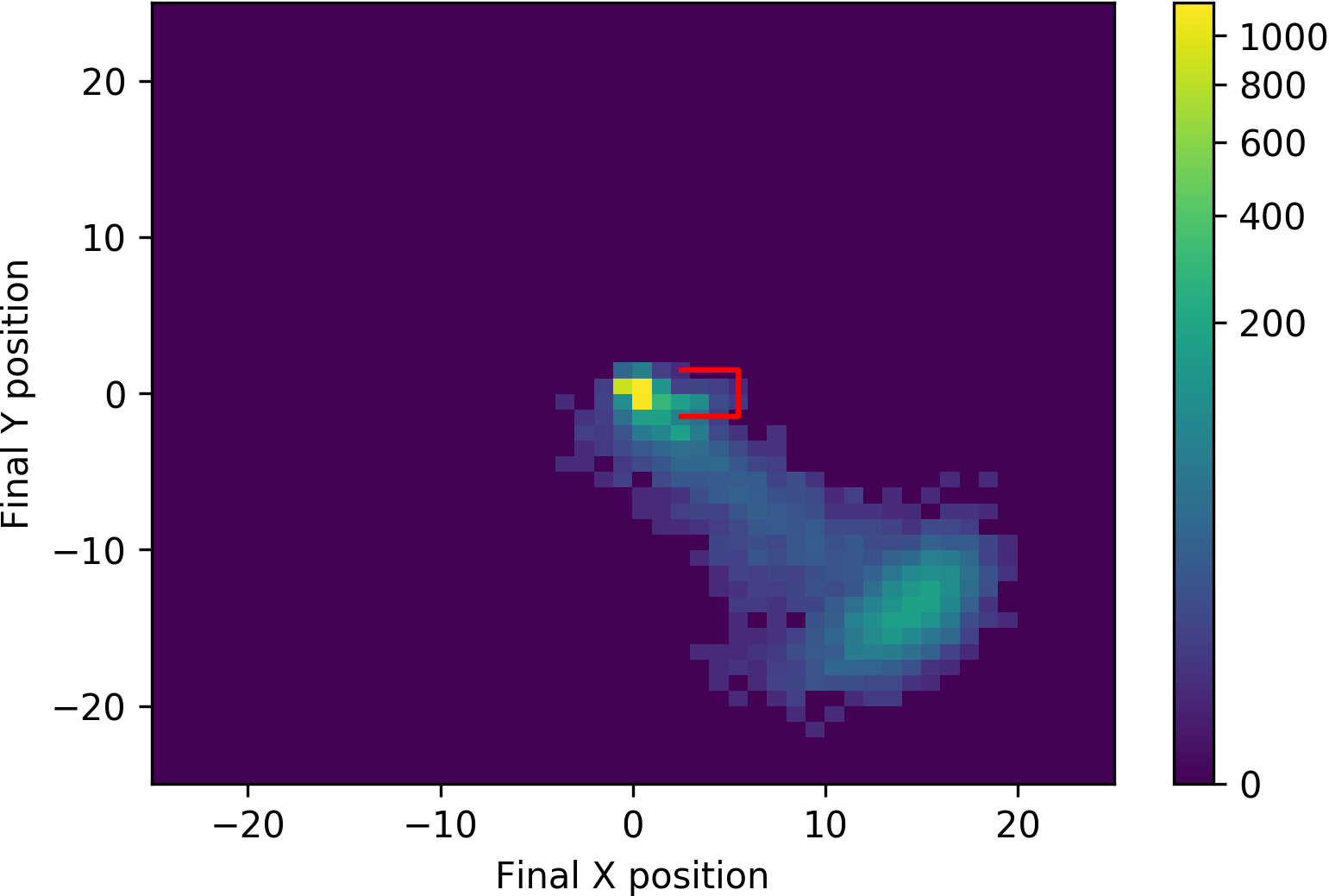}
    }
    \end{subfigure}
    \caption{2D histogram of the final positions of the population from the last generation for the Deceptive Ant and Deceptive Humanoid tasks. The red lines represent the walls of the trap, which makes the problem deceptive. For both environments, ES ends up trapped, (a) and (c), while QE-ES is able to overcome the local optimum, (b) and (d). Please note the change in scale.}
    \label{fig:exp3_histograms}
\end{figure}

\subsubsection{Task 3: Deceptive Locomotion}
The final experiment was done on the deceptive variant of the environment, to test which algorithm can deal with deceptive problems (see Fig \ref{fig:exp3_curves} and Fig \ref{fig:exp3_histograms}). Both with the Deceptive Ant and Deceptive Humanoid environments, ES failed to escape the trap, a result consistent with previous work \citep{conti2017improving}. QE-ES had no problem escaping the trap in both experiments. This result demonstrates that Quality Evolvability ES is able to cope with at least some level of deceptiveness, even though it does not have the ever increasing pressure to escape like Quality Diversity algorithms.

\section{Conclusion}



We have presented Quality Evolvability ES, a technique to simultaneously select for evolvability and fitness. QE-ES allows us to benefit from evolvability in cases when evolvability and fitness are not aligned. 
While our experiments demonstrated that evolvability can increase the performance of evolution in a single environment, it is yet to be determined whether the learned evolvability is general enough to be useful in different environments and with different behaviour characterizations.
Quality Evolvability and Quality Diversity are not mutually exclusive approaches. In future work, we aim to explore the direction of looking for an archive of diverse and evolvable individuals.

\section{Acknowledgements}

This work was supported by the EPSRC Centre for Doctoral Training in Intelligent Games \& Game Intelligence (IGGI) [EP/L015846/1].

\footnotesize
\bibliographystyle{apalike}
\bibliography{example} 

\end{document}